\documentclass[final]{cvpr}

\usepackage{times}
\usepackage{epsfig}
\usepackage{graphicx}
\usepackage{amsmath}
\usepackage{amssymb}
\usepackage{color}
\usepackage{enumitem}

\usepackage[pagebackref=true,breaklinks=true,colorlinks,bookmarks=false]{hyperref}

\def\cvprPaperID{9} 
\def\confYear{CVPR 2021 AICCW}

\newcommand{\subheading}[1]{\textbf{#1}}

\newif\ifsubmit
\submitfalse
\ifsubmit
    \newcommand{\nate}[1]{}
    \newcommand{\peggy}[1]{}
    \newcommand{\irfan}[1]{}
\else
    \newcommand{\nate}[1]{{\textbf{\color{red}Nate:{#1}}}}
    \newcommand{\peggy}[1]{{\textbf{\color{red}Peggy:{#1}}}}
    \newcommand{\irfan}[1]{{\textbf{\color{red}{Irfan:{#1}}}}}
\fi

\begin{document}

\title{Automatic Non-Linear Video Editing Transfer}

\author{Nathan Frey, Peggy Chi, Weilong Yang, and Irfan Essa\\
Google Research\\
1600 Amphitheatre Pkwy, Mountain View, CA 94043\\
{\tt\small nle-transfer@google.com}
}

\maketitle

\begin{abstract}
We propose an automatic approach that extracts editing styles in a source video and applies the edits to matched footage for video creation. Our Computer Vision based techniques considers framing, content type, playback speed, and lighting of each input video segment. By applying a combination of these features, we demonstrate an effective method that automatically transfers the visual and temporal styles from professionally edited videos to unseen raw footage. We evaluated our approach with real-world videos that contained a total of 3872 video shots of a variety of editing styles, including different subjects, camera motions, and lighting. We reported feedback from survey participants who reviewed a set of our results.
\end{abstract}

\section{Introduction}

The easy access to digital cameras and editing tools has enabled users to vastly generate online video content. From sports, music, to dialogue-driven videos, content creators derive and apply a variety of video editing styles~\cite{ViralVideoStyle_2014}, such as short cuts~\cite{DialogueDriven}, moment highlights~\cite{BasketballHighlights_2016} or artistic looks~\cite{StylizingVideobyExample_2019}.
Effective editing styles can engage the audience in video consumption~\cite{arijon1991grammar}. However, video editing is a time- and effort-consuming process. It requires composing raw footage onto a timeline and continuously applying both visual and temporal effects, known as \emph{``non-linear video editing''}~\cite{arijon1991grammar}. An editor makes a series of careful decisions to consider the subjects, camera motion, and visual quality in each scene, which can be especially challenging to produce multi-shot videos with consistent styles.

Recent research has demonstrated advanced methods of style transfer for \emph{images} that control fine-grained attributes, such as spatial and color information~\cite{NeuralStyleTransfer_2017} or semantic structure~\cite{AttributeTransfer_2017}. The performance can be in real-time~\cite{NeuralStyleTransferStability_2017} with unconstrained styles~\cite{10.5555/3294771.3294808} for photorealistic images~\cite{PhotorealisticStyleTransfer_2020}.
To apply visual styles to a \emph{video with temporal effects}, researchers have proposed methods to preserve the visual quality from example videos~\cite{StylizingVideobyExample_2019,10.1145/3386569.3392453} and consider temporal coherence~\cite{CoherentOnlineVideoStyleTransfer_2017,10.1145/3386569.3392457}.
%
In addition, automatic techniques can edit domain specific videos such as interviews~\cite{InterviewVideo}, dialogues~\cite{DialogueDriven,transcript_2019}, or instructional content~\cite{DemoCut_UIST13,QuickCut} and can leverage metadata~\cite{MetadataVideoEditing_DIS02}. 
However, there is limited prior work on transferring personal editing styles, especially for creators who have produced a set of example videos.





\begin{figure}[t!]
\centerline{\includegraphics[trim=102 30 65 20,clip,width=\columnwidth]{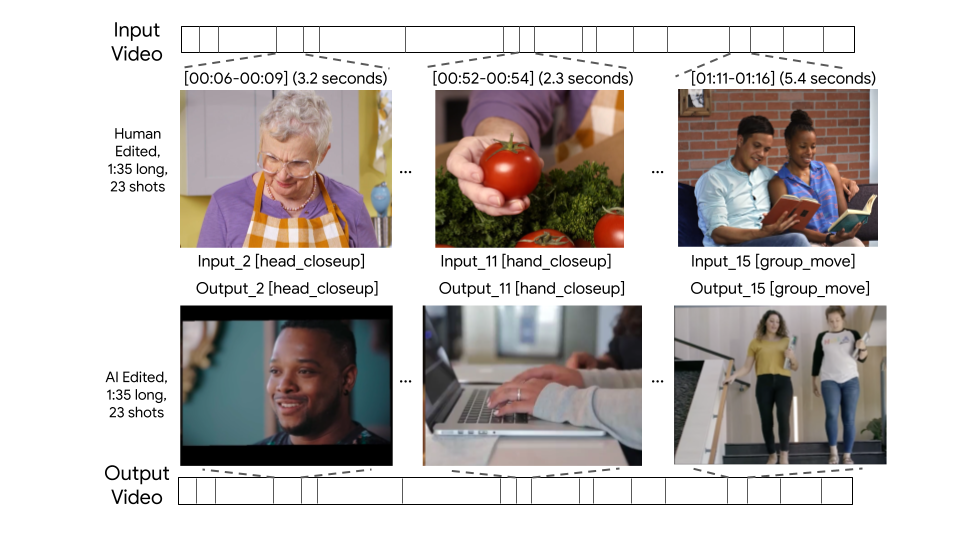}}
\caption{Given an example edited video, our automatic approach extracts the editing styles in each video segment, selects a matched footage from a repository, and applies the similar style to compose an output video. The top row shows the annotation of the source video generated by our pipeline, where blue keypoints label background objects and red keypoints denote human foreground objects in a video frame.}
\label{fig:teaser}
\end{figure}

In this paper, we introduce an automatic approach for transferring the editing style from a source video to a target footage repository (see Figure~\ref{fig:teaser}). Given an edited input video, our Computer Vision based techniques perform shot detection and analyze each shot's editing style, including framing, content type, playback speed, and lighting. We then apply a combination of these visual and temporal styles to unseen raw footage in a user-provided repository. We tested our pipeline with professional-edited videos and reported feedback from survey participants.
Specifically, our work makes the following contributions: 
\begin{itemize}[leftmargin=*]
    \item An automatic approach to extract editing styles from an edited video and transfer to new raw footage.
    \item Methods to track camera motion, scene types, and lighting in an edited video.
    \item An informal evaluation of automatically edited videos with transferred styles.
\end{itemize}



\section{Background: Video Editing Techniques}
We build our work upon conventional video editing techniques that create effects on both visual and temporal domains of a video~\cite{arijon1991grammar,FilmArt_2013}. Video editors commonly combine multiple techniques to develop a unique editing style. Specifically, we focus on a subset of fundamental effects:

\subheading{Framing} presents visual elements by considering the spatial placement of one or more subjects, which guide the audience's attention. Framing a series of video frames leads to the \emph{camera motion}, which is a core element of filmmaking.
A cinematic camera can move in multiple ways, including stationary and Pan–Tilt–Zoom (PTZ) that are commonly seen in user-generated videos. While unrestricted camera mounts allow for full six degrees of freedom (6DoF) motion, they are beyond our scope given the challenges of reproducing parallax effects on raw footage.
%

\subheading{Transition} is a mechanism to connect video shots temporally.
Common techniques include hard cuts (i.e., direct shot changes) and lighting-based transitions (e.g., fades) that we utilize in this work.  
%
Transitions often consider the connection of content and framing between shots. Placements of subjects across multiple shots may create surprising effects, such as a ``match cut`` that intentionally aligns shapes, colors, and movements of two distinct shots~\cite{FilmArt_2013}.

\subheading{Playback speed} of a video segment adjusts the narrative pace, from slow (less than 1x), normal (1x), to fast (greater than 1x). Manipulating the speed often informs the importance or the emotion of a subject. For example, slow motion increases the drama of footage while high speed can increase the humor; speed ramps can enhance specific moments of a continuous action. 

\subheading{Color and lighting} change the tone and mood of a video frame and can be adjusted during the capturing or the post-production phase. It is common that scenes in an edited video have applied color filters and lighting adjustments to enhance storytelling. 


\begin{figure}[t!]
\centerline{\includegraphics[trim=25 20 30 50,clip,width=\columnwidth]{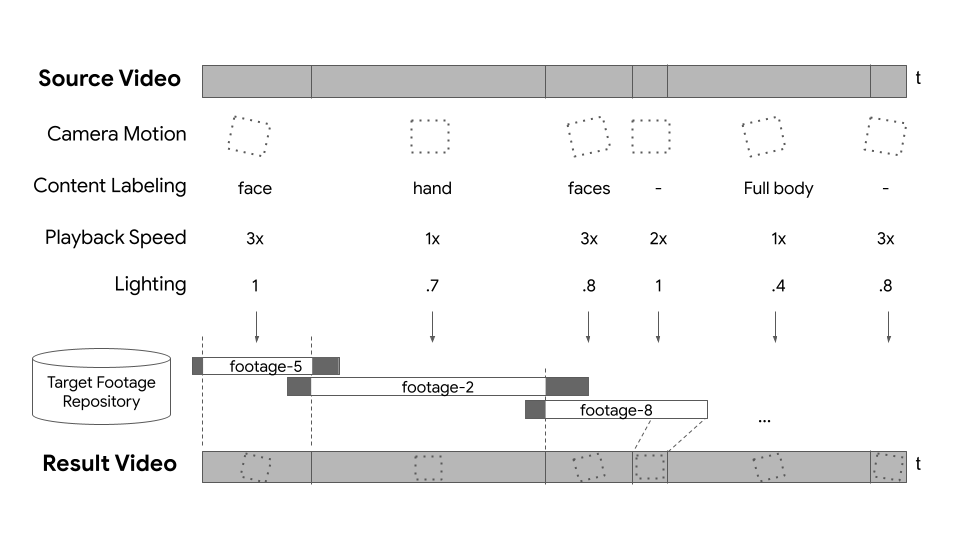}}
\caption{Our pipeline automatically segments an edited video and detects the detailed editing decisions, including camera motion, content, playback speed, and lighting. It iteratively selects a matched footage from a repository and applies the similar style.}
\label{fig:pipeline}
\end{figure}

\section{Automatic Non-Linear Video Editing}

We present an automatic approach for non-linear video editing that transfers the styles from a source edited video to new target footage from a repository (see Figure~\ref{fig:pipeline}). We focus on video footage that has limited or no on-screen audio, which is commonly seen in a music video with a sound track, short form content such as a TikTok video or a documentary video with a voiceover.
%
Our pipeline automatically performs shot detection on a source video. Each video shot is then processed independently for which transitions, camera motion, lighting effects, playback speed and content type detected and stored.
Based on the extracted editing styles and criteria, we search for the most appropriate raw footage and apply the style to compose an output video. 

\subsection{Shot Detection}
A video shot is a continuous sequence of video frames without a cut or a jump~\cite{arijon1991grammar}. To detect the occurrence of a shot change, we compute the color histogram of frames, which is a common technique and can handle gradual transitions in edited videos~\cite{MakeupBreakdown_2021}. 
%
Next, we group similar shots into a scene based on FAST keypoint matching~\cite{FAST_2010}. To determine visual similarity, we extract keypoints and descriptors from a set of representative frames, efficiently match keypoint sets across frames using an approximate nearest-neighbor search. We apply weak geometric constraints in the form of the fundamental matrix to arrive at a set of spatially consistent keypoint matches.

\subsection{Scene Content Labeling}
For each detected scene, we sample multiple frames and label the content to identify the focused subjects. We perform object detection on 100 labels, including faces, human poses, and hands using MobileNets ~\cite{MobileNets_2017s}.
We focus on the following content categories:
(1) We label a scene that consists of a small number of subjects or objects as a \emph{single focus scene}. Editors commonly design the placement of an object and camera motion around the object to direct the audience's attention. 
(2) A \emph{multi-subject scene} consists of a medium number of objects, such as a group of people or products. Typically, the distribution of the subjects guides the audience. The camera motion is weighted towards its overview rather than following a single subject.
%
(3) A \emph{background scene} has little or no specific object focus, such as presenting a landscape, a crowd, and general b-roll footage. The camera motion is likely not related to any specific point in the footage. 
%


\begin{figure}[t!]
\centerline{\includegraphics[trim=100 180 180 50,clip,width=0.95\columnwidth]{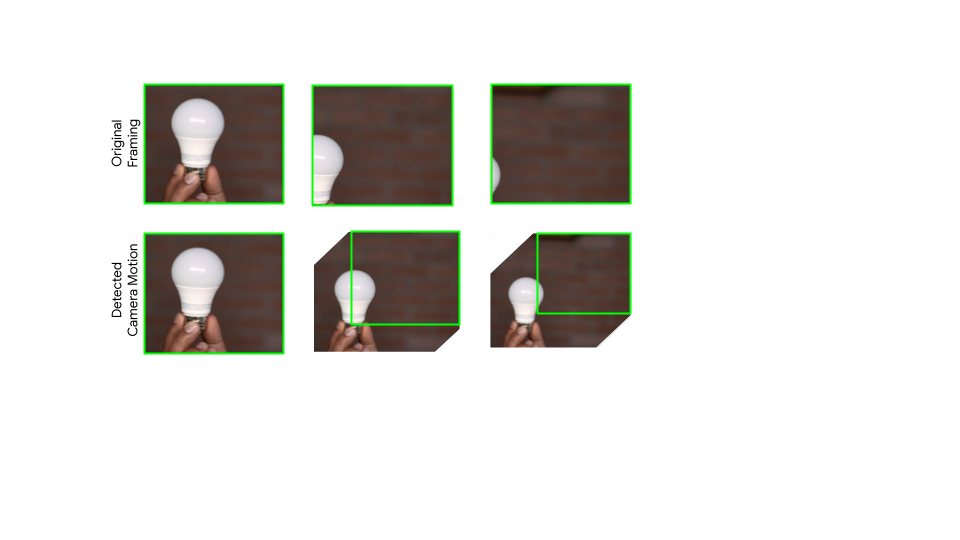}}
\caption{Our pipeline performs keypoint tracking per video frame to form the homography (top) and captures the camera motion, visualized as mosaic over time from the start location (bottom).}
\label{fig:motion}
\end{figure}

\subsection{Editing Style Extraction}

\subsubsection{Camera Motion}
We represent the camera motion as a reprojection of video frames, denoted as a projective transformation $H_{ref}$.
Our method captures the camera motion through frame-to-frame keypoint matching with a RANSAC based homography solver~\cite{RANSAC_1981,CrossCameraTracking_2013} (see Figure~\ref{fig:motion}). We reject keypoints on a face or body to remove foreground objects from the homography estimation of the camera's actual motion.
The total reprojection of a frame at time $t$ is the left-multiplied accumulation of prior frame's projections:
\[ H_{ref}(t) = H_{ref}(t)*H_{ref}(t-1)*H_{ref}(t-2)...*H_{ref}(t^{start}) \]


\subsubsection{Playback Speed and Brightness}
While Computer Vision techniques can classify motions in a video as fast, normal, or slow, we rely on users to label the accurate playback speed from a source video as a constant value (e.g. 0.3x). To capture the brightness, we implemented a brightness detector to compute the average value of the pixels in a scene.

\subsection{Style Transfer}
Based on the editing attributes detected from the source video, we extract features from all raw footage in the target repository. We then select matched footage and transfer the edit styles to compose an output video, rendered using the open-source framework MediaPipe~\cite{MediaPipe}.

\subsubsection{Footage Retrieval}
For each shot from the source video, we retrieve a matched raw footage clip from the target repository by style criteria: First, we enforce hard constraints, including the aspect ratio and duration after playback speed resampling. Next, we filter by content type to select similar target scenes. Specifically, content scene labels (e.g., single focus shot) are used to match a scene by their overall sentiment. Finally, we prioritize by similar objects to best match the input shot.

\begin{figure}[t!]
\centerline{\includegraphics[trim=80 160 180 50,clip,width=0.95\columnwidth]{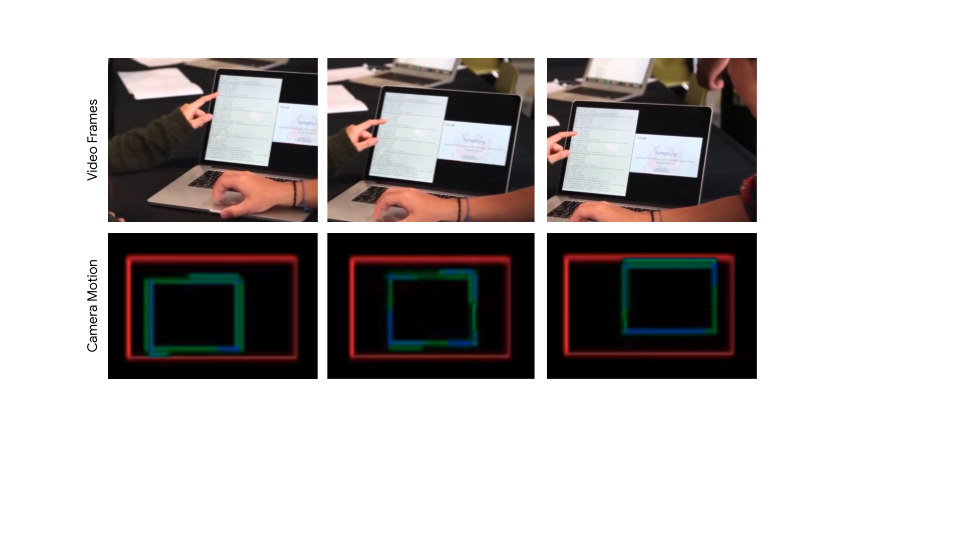}}
\caption{The camera motion from the source video (top) is positioned within the target footage (bottom red) in order to frame the subjects within the viewpoint and maximize the size and coverage of scene content.}
\label{fig:motion_apply}
\end{figure}

\subsubsection{Camera Motion}

To apply frame-to-frame camera motion from the source video, we first stabilize the target footage to remove its camera motion $H_{target}^{-1}(t)$.
The final reprojection on each frame is the combination of undoing the original projection on the target footage and applying the camera motion from the source footage:
\[ H_{target}^{-1}(t) \times H^{ref}(t)\]

We apply framing for each shot with a starting position, defined as an initial scale and offset
$H_{ref}(t=start)$.
For a \emph{single focus shot}, we align this start reprojection with the target content, while enforcing the full duration of the camera motion within the bounds of the target footage.  If the reference footage zooms out, this start position starts zooming in within the bounds of the video frame (see Figure~\ref{fig:motion_apply}).
%
For a \emph{general scene}, the camera motion maximises the size in which it repositions in the frame via the same process.

\subsubsection{Playback Speed and Brightness}
Frame resampling is used to transfer the playback speed from source to target before reprojection:
\[ H_{target}^{-1}(s*t) * H_{ref}(s*t)\]

\begin{figure}[t!]
\centerline{\includegraphics[trim=140 250 150 30,clip,width=0.9\columnwidth]{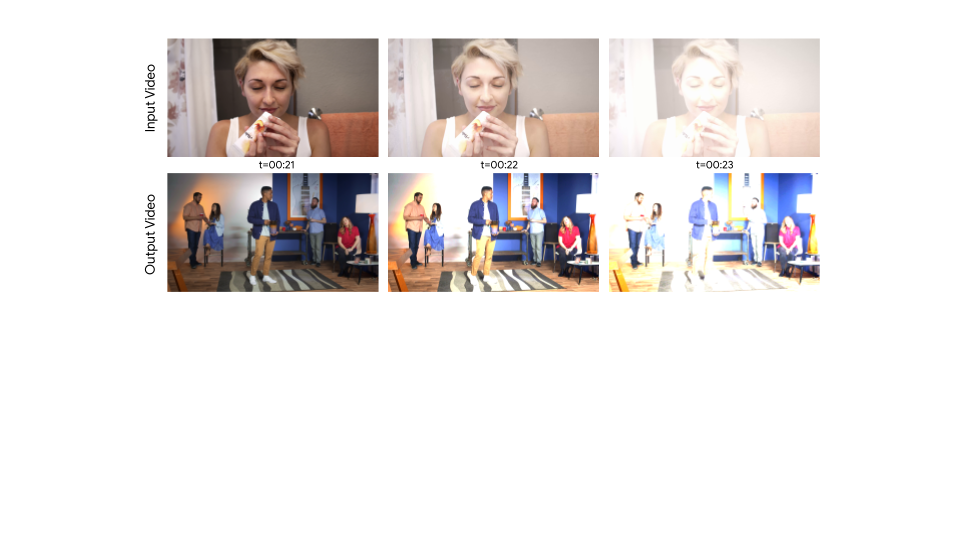}}
\caption{The brightness of the source edited video (top) is transferred to the target footage (bottom) as the video fades through black for a scene change.}
\label{fig:lighting}
\end{figure}

We then apply the per-frame value of stylistic lighting adjustments from the source video to target footage (see Figure~\ref{fig:lighting} for an example of fade-through-black transition).

\newcommand{\numberOfProcessedVideos}[1] {60}
\newcommand{\numberOfProcessedVideoShots}[1] {3,872}
\newcommand{\numberOfSurveyResponses}[1] {15}
\newcommand{\numberOfSurveyResponsesToVideos}[1] {105}
\newcommand{\userquote}[1]{``\emph{#1}''}

\begin{figure}[t!]
\centerline{\includegraphics[trim=140 60 150 60,clip,width=0.88\columnwidth]{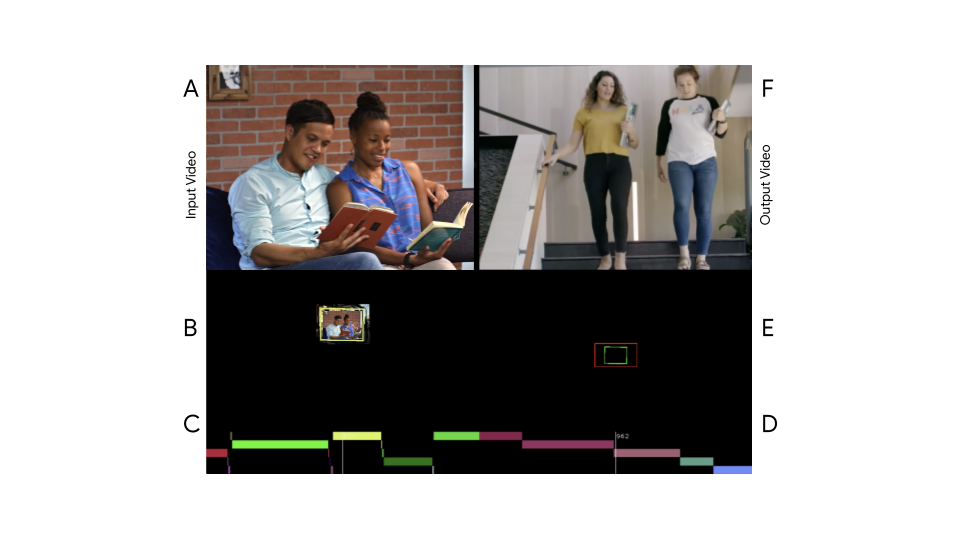}}
\caption{We visualize editing results in a review panel that include (A) the original source video annotated by keypoints, (B) the mosaic form of the captured camera motion, (C-D) the synchronized video timelines of the source and target videos showing shot changes, (E) subcropping of target footage that matches the camera motion of the mosaic, and (F) the output video.}
\label{fig:review}
\end{figure}

\section{Experiments}


We selected \numberOfProcessedVideos{} professional-edited videos from multiple creators on YouTube as the source videos to test our pipeline. We built our test set based on the following criteria: a video (1) includes more than 10 shots showing different subjects, (2) contains a mixture of camera views, and (3) has independent sound track.
The source videos are 2-minute long on average.
The repository of target footage contains 40 raw videos, ranging from 30 to 90 seconds long.

\subsection{Results}

\begin{figure}[t!]
\centerline{\includegraphics[trim=100 50 180 30,clip,width=0.95\columnwidth]{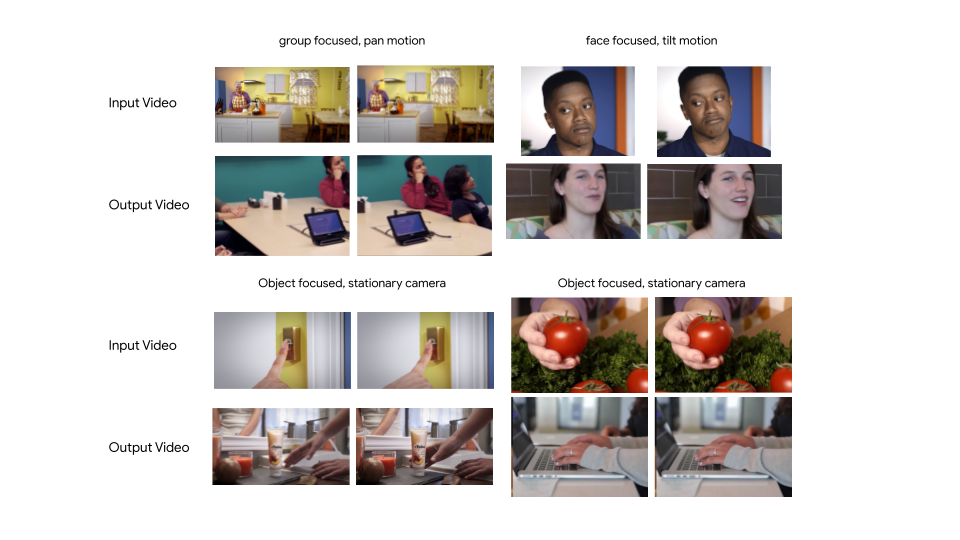}}
\caption{Example results from our pipeline.}
\label{fig:reuslts}
\end{figure}

Table~\ref{tab:result_analysis} presents the analysis of our results on \numberOfProcessedVideos{} videos. Figure~\ref{fig:reuslts} shows our example videos.
We reviewed the results using a visualization UI (see Figure~\ref{fig:review}).
Our pipeline detected a total of \numberOfProcessedVideoShots{} shots that include both long and short video shots of various content types and visual tones.
We observed that the pipeline especially well captured camera motions, which can be difficult for average creators to edit and transfer manually without assistance. In addition, the pipeline selected and positioned subjects from raw footage similar to source videos with the exact temporal allocation. The combination of these editing decisions therefore provides a similar style.


\subsection{Audience Feedback}

To gather audience feedback, we conducted an online survey within our organization. Each survey participant was asked to review 7 selected video segments, each was 10 to 20 seconds long and contained 3-6 shots. We synchronized the source and output videos, presented side-by-side as one video for review. Each video was followed by three Likert-scale questions on the matches of \emph{editing style}, \emph{camera motion}, and \emph{visual quality} between the source and output videos. We also collected text comments.

We received \numberOfSurveyResponses{} responses with a total of \numberOfSurveyResponsesToVideos{} ratings to the videos per question. Overall, participants agreed that the editing style between the source and output videos is similar ($Median=4$ overall and for six of the seven videos), as well as the camera motion ($M=4$ overall and for all videos). They commented, \userquote{it's (the result video is) almost as good as the left one (the source video).} (on Test Video \# 4) and \userquote{very consistent motion} (on Test Video \# 2).
We also learned that when comparing styles, people paid attention to the subjects, resolution, and quality carefully, while we received a neutral rating to the similarity of visual quality ($M=3$). One participant commented, \userquote{right video has some visible artifacts like video slowdown which was obviously filmed in lower fps — this completely breaks the immersion.} (on Test Video \# 3), and another noted, \userquote{resolution and camera focus on right video is inconsistent.} (on Test Video \# 5).
These findings inspire us to focus effort on quality understanding, super resolution, and storytelling that we are currently developing.


\begin{table}[t!]
\caption{We performed style extraction on 60 online videos, resulting in \numberOfProcessedVideoShots{} shots with a variety of content and motions.}
\centerline{\includegraphics[trim=200 80 200 60,clip,width=.99\columnwidth]{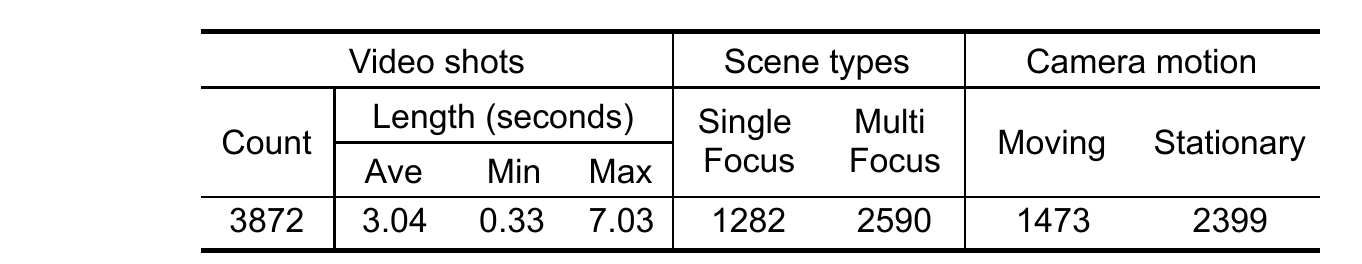}} 
\label{tab:result_analysis}
\end{table}

\section{Conclusion}

We present an automatic approach that extracts video editing styles from a source video and applies the edits to raw footage in a repository. Our technique detects shots and content in an edited video. For each shot, it analyzes the framing, playback speed, and lighting as the editing choices. By applying a combination of these features, our method transfers both visual and temporal styles from an edited video to unseen footage. Our preliminary results and feedback from survey participants show that editing styles can be automatically transferred with reasonable effects.

{\small
\bibliographystyle{ieee_fullname}
\bibliography{egbib}
}

\end{document}